\documentclass[letterpaper, 10 pt, conference]{ieeeconf}  

\IEEEoverridecommandlockouts                              

\usepackage{times}
\usepackage{xcolor}
\usepackage[linesnumbered,ruled,vlined]{algorithm2e}

\usepackage{multicol}
\usepackage[bookmarks=true]{hyperref}

\usepackage{bm}
\usepackage{amsmath} 
\usepackage{amssymb}  
\usepackage{amsfonts}
\usepackage{mathrsfs} 

\usepackage{subfigure}
\usepackage{graphicx}
\usepackage{gensymb} 

\usepackage{booktabs}
\usepackage{multirow}
\usepackage{tablefootnote}
\usepackage{threeparttable}

\def\calL{\mathcal{L}}

\def\Rbb{\mathbb{R}}
\def\Rbb3{\mathbb{R}^3}

\newcommand{\myparagraph}[1]{\noindent{\bf #1}}

\title{\LARGE \bf
Gaussian Semantic Field for One-shot LiDAR Global Localization
}

\author{Anonymous Submission}

\author{Pengyu Yin, Shenghai Yuan, Haozhi Cao, Xingyu Ji, Ruofei Bai, Siyu Chen, and Lihua Xie, \emph{Fellow, IEEE}}

\makeatletter
\renewcommand{\p@table}{}
\makeatother

\begin{document}

\maketitle
\thispagestyle{empty}
\pagestyle{empty}

\begin{abstract}

We present a one-shot LiDAR global localization algorithm featuring semantic disambiguation ability based on a lightweight tri-layered scene graph. While landmark semantic registration-based methods have shown promising performance improvements in global localization compared with geometric-only methods, landmarks can be repetitive and misleading for correspondence establishment. We propose to mitigate this problem by modeling semantic distributions with continuous functions learned from a population of Gaussian processes. Compared with discrete semantic labels, the continuous functions capture finer-grained geo-semantic information and also provide more detailed metric information for correspondence establishment.
We insert this continuous function as the middle layer between the object layer and the metric-semantic layer, forming a tri-layered 3D scene graph, serving as a light-weight yet performant backend for one-shot localization. We term our global localization pipeline Outram-GSF (Gaussian semantic field) and conduct a wide range of experiments on publicly available data sets, validating the superior performance against the current state-of-the-art. 
\end{abstract}

\section{Introduction}

First proposed in \cite{armeni20193d}, the 3D scene graph has gained increasing attention over the years. A 3D scene graph is a multilayered, compact representation of an environment that organizes spatial-semantic information into interconnected layers, typically comprising metric-semantic elements at the lowest level, objects at intermediate levels, and places or rooms at higher levels. This hierarchical structure captures not only the geometric and semantic properties of individual entities but also their topological and spatial relationships, enabling efficient reasoning about scene structure. Given its rich semantic information and compactness, several recent works \cite{wang2025sgt, yin2024outram, wang2024sglc, ma2024tripletloc} leverage it as a tool for large-scale outdoor global localization. 




In the literature, Outram \cite{yin2024outram} proposed to solve the LiDAR global localization problem in a registration manner, where a double-layered 3D scene graph is used for a coarse-to-fine hierarchical search. The core idea behind it is that the query point cloud can be parametrized as lightweight instance-level triangle descriptors. And the global localization can be achieved in a one-shot manner by a coarse-to-fine group-to-individual search. Subsequent variants \cite{wang2024sglc, wang2025sgt, huang2025sgtd, ma2024tripletloc} focus on designing different descriptor generation manners for better discrimination or efficiency. 
Nevertheless, these methods rely solely on the centroids of semantic clusters and their topological connections, overlooking the rich spatial-semantic distributions within and between clusters. While centroids provide coarse localization anchors, they fail to capture the detailed geometric variations and semantic gradients that characterize real-world environments. The continuous spatial distribution of semantic information of how semantics evolve across space has the potential to provide fine-grained cues for robust correspondence establishment. 
To this end, we propose a novel global localization approach that integrates Gaussian semantic fields into 3D scene graphs. Unlike discrete semantic labels, Gaussian semantic fields (GSF) provide continuous semantic-geometric representations that capture fine-grained environmental semantic information. By incorporating GSF as the middle layer of a 3D scene graph, we develop a one-shot global localization system with superior robustness against semantic repetitiveness.

\begin{figure}[t]
\label{fig: geo_symmetric}
\centering
\includegraphics[width=\columnwidth]{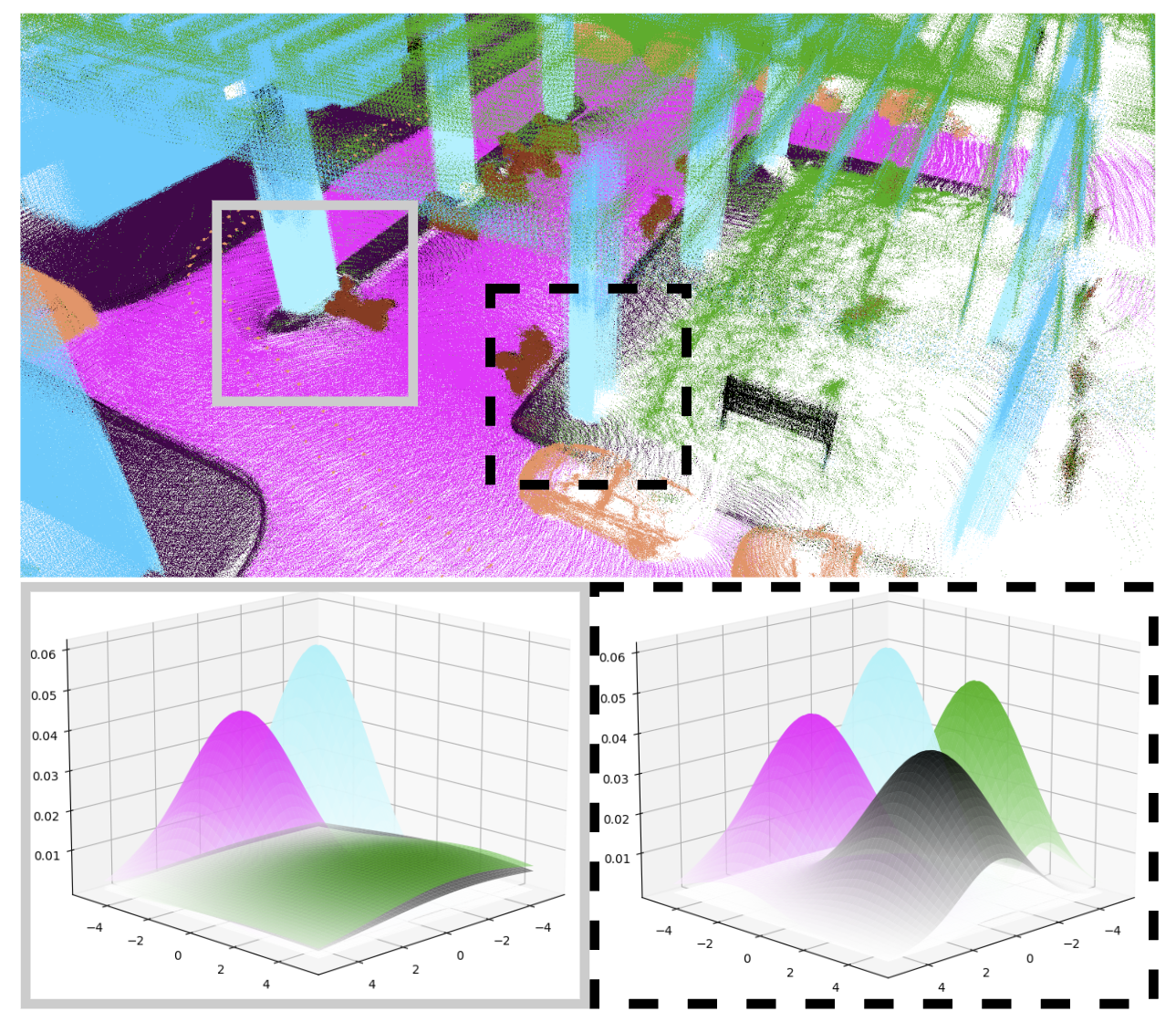}
\caption{A semantic annotated point cloud of a partially semantically ambiguous scene and two Gaussian semantic fields of the corresponding rectangle areas. The semantic field encodes the metric-semantic distribution of a small area (purple for road, light blue for pole, green for vegetation, and black for traffic sign). Note that the two areas are highly symmetric in geometry, even in instance-level semantics, but naturally differentiable in surrounding vegetation distributions. We leverage this continuous semantic distribution for robust global localization.}
\end{figure}



To summarize, we present Outram-GSF, a one-shot global LiDAR localization framework that addresses semantic ambiguity through continuous spatial-semantic modeling. Our main contributions are:
\begin{itemize}
\item We introduce Gaussian Semantic Fields (GSF) as a plug-in, intermediate layer in 3D scene graphs, enabling continuous modeling of spatial-semantic distributions.

\item We develop a probabilistic framework based on Gaussian processes for learning semantic distributions from local regions, along with a principled similarity metric that leverages these continuous representations for robust correspondence establishment.
\item We demonstrate through extensive experiments on public datasets that our approach achieves state-of-the-art performance in one-shot global localization, particularly excelling in scenarios with repetitive semantic structures.
\end{itemize}

\section{Literature Review}

\myparagraph{One-shot Semantic LiDAR Localization.} Contrary to many submap-based LiDAR global localization methods \cite{dube2017segmatch, dube2020segmap, oliveira2025regrace} that require the mobile robot to move for point cloud accumulation, one-shot methods aim at performing localization with only one current LiDAR scan. These methods are achieved by either bag-of-words (BoW) query-based methods \cite{yuan2023std, yuan2024btc} or registration-based methods \cite{ankenbauer2023global}. BoW-based methods are usually extensions of loop closure detection pipelines, as the same query-based nature can be shared. However, these per-scan-based methods all share discretization issues where the query scan can never exist in the map database. Recent advances in point cloud semantic scene understanding have motivated the integration of semantic information into one-shot semantic LiDAR localization methods. By leveraging semantic information, these approaches can construct 3D scene graphs \cite{armeni20193d} that provide a hierarchical and abstract representation of the environment, resulting in significantly more lightweight maps compared to dense point clouds. Moreover, scene graphs enable registration-based methods at the object level, where robust correspondences can be established between semantic entities rather than raw geometric features. The encoded instance-wise topological relationships provide rich contextual constraints that significantly reduce the search space for correspondence inference, enabling more efficient and robust localization. Outram \cite{yin2024outram} proposed to parametrize the semantic annotated query point cloud can be parametrized as lightweight instance-level triangle descriptors. And the global localization can be achieved in a one-shot manner by a coarse-to-fine group-to-individual search against the entire map. Subsequent variants \cite{wang2024sglc, wang2025sgt, huang2025sgtd, ma2024tripletloc} focus on designing different descriptor generation manners exploiting instance-wise topological information for better discrimination or efficiency. Nevertheless, they all depend exclusively on the centroids of semantic instances and the instance-wise topology, facing a severe performance drop in semantic ambiguity environments (as shown in Fig. (\ref{fig: geo_symmetric})). As such, the instance-only semantic information can never establish the correct correspondences. 

\myparagraph{Spatial Semantic Modeling} Early spatial semantic modeling approaches discretize 3D space into voxels or segments with categorical labels \cite{behley2019semantickitti}, treating semantic properties as spatially independent discrete variables. Recent advances have shifted toward continuous spatial-semantic representations that model the spatial distribution and correlation of semantic information, jointly encoding geometric and semantic properties \cite{gan2020bayesian, ghaffari2018gaussian}. Gaussian processes have been successfully applied to model spatial semantic distributions \cite{o2012gaussian, ghaffari2018gaussian}, providing probabilistic frameworks that capture both local semantic variations and global spatial correlations. These methods leverage spatial kernels to encode the assumption that semantically similar regions exhibit spatial continuity \cite{kim2013continuous}. Hilbert maps \cite{ramos2016hilbert} and Bayesian generalized kernel inference \cite{gan2020bayesian} further extend this concept by learning discriminative models for semantic occupancy in continuous space. More recently, implicit neural representations such as semantic neural fields \cite{zhi2021place, vora2021nesf} have demonstrated the ability to encode fine-grained spatial-semantic information in a continuous, differentiable manner. However, most existing approaches focus on dense scene reconstruction rather than efficient localization. The challenge remains in extracting compact yet informative spatial-semantic representations that preserve both local geometric details and global semantic context for robust correspondence establishment \cite{chang2023hydra}. 

The limitations of existing semantic localization approaches, coupled with recent advances in spatial semantic modeling, motivate the integration of fine-grained semantic information into the localization pipeline. Our work addresses this gap by introducing Gaussian semantic fields (GSF) as an intermediate layer in 3D scene graphs, enabling continuous spatial-semantic modeling while maintaining the computational efficiency required for global localization. By capturing detailed geo-semantic distributions within local regions, GSF significantly enhances correspondence establishment in semantically ambiguous environments where traditional discrete labels fail to provide sufficient discriminative power.

\section{Methodology}
\begin{figure*}[t]
\centering
\includegraphics[width=\textwidth]{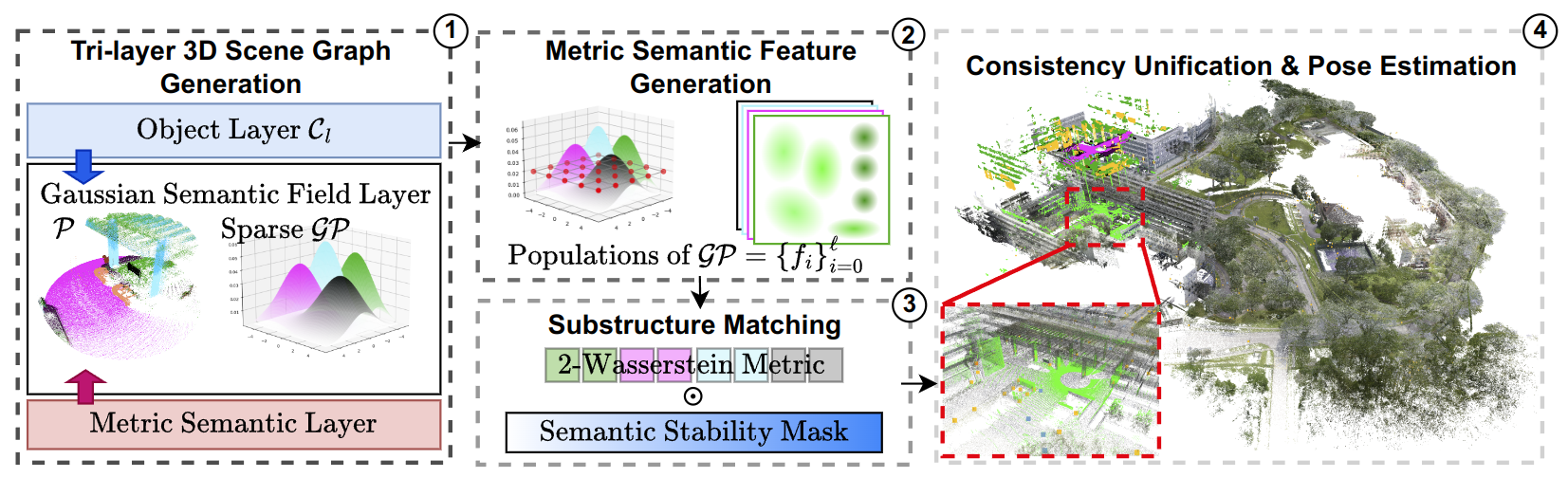}
\caption{We present Outram-GSF, a one-shot global LiDAR localization pipeline. In stage 1, we generate tri-layer scene graphs for both the current query scan and the target reference map. Around each instance in the object layer $\mathcal{C}_1$, we train sparse Gaussian processes (GPs) using the semantic annotated point clouds in the metric semantic layer. Populations of GPs for metric similarity comparison are generated by linear probing in stage 2. Combined with a semantic stability mask, the populations of GP can be used as a similarity measurement in the Wasserstein sense in stage 3. Given the correspondences established by the above process, inlier correspondences are selected by maximum clique (stage 4).}
\label{fig: teaser_fig}
\end{figure*}

We formulate global localization as a global point cloud registration problem between the query LiDAR scan $\mathcal{P}=\left\{\mathbf{p}_i\in\mathbb{R}^3\right\}_{i=1}^{n}$ and the reference point cloud map $\mathcal{M}=\left\{\mathbf{m}_j\in\mathbb{R}^3\right\}_{j=1}^{m}$. The optimal rigid transformation $\mathbf{T} \triangleq \left[\mathbf{R},\mathbf{t}\right] \in \text{SO(3)} \times \mathbb{R}^3$, including the rotation $\mathbf{R}$ and translation $\mathbf{t}$, can be estimated in close-form \cite{pomerleau2015review} given the ground truth correspondence set $\mathcal{I}^{\star} = \left\{\left(i,j\right)\right\} \in[n]\times[m]:=\left\{1,\dots,n\right\}\times\left\{1,\dots,m\right\}$. Directly applying point cloud registration techniques to global localization faces two challenges: 1) A proper representation of both query scan and reference map for scalability; 2) An extremely robust correspondence establishment strategy for inlier selection. To address the former challenge, we adhere to the pipeline presented in the previous work \cite{yin2024outram} to use a compact 3D scene graph as the scene representation. For the second challenge, a semantic field layer is further injected for better correspondence setup in semantic ambiguous environments.

\subsection{Overview}
The proposed global localization pipeline includes 4 stages: 1) A tri-layer 3D scene graph generation module to produce the Gaussian semantic field layer between the upper object layer and lower semantic point cloud layer from a sparse Gaussian process; 2) A set of metric semantic features is generated by grid probing approach; 3) A graph-based substructure matching back-end that generates instance-level correspondences; 4) A consistency unification process to generate inlier correspondences and produce the final pose estimation.

\subsection{3D Scene Graph with Local Gaussian Semantic Field}
\label{sec: GSF gen}
The 3D scene graph is a multi-layered, hierarchical representation for joint modeling of low-level geometries and high-level, human-perceivable semantics. We inject another layer, the Gaussian semantic field layer, between the object layer and metric semantic point layer in the commonly used 3D scene graph structure \cite{strader2024indoor}. The Gaussian semantic field layer is created by training a sparse Gaussian process $\mathcal{GP}$. This layer is designed to fully capture the continuous semantic-geometric distribution and provide more holistic semantic information for global localization.

We first build a typical double-layered 3D scene graph for both the current query scan and the reference map by semantic segmentation \cite{cao2024mopa} and instance clustering. Given a set of points $\mathcal{P}=\left\{\mathbf{x}_i\right\}_{i=1}^{n}$, we define a mapping function $\lambda$ from an off-the-shelf segmentation network \cite{cao2023multi}: $\left\{l_i, \mathbf{y}_i \right\}= \lambda(\mathbf{x}_i)$, where $l_i \in \calL \subset \mathbb{N}$ is a finite set of semantic labels and $\mathbf{y}_i \in \mathbb{R}^{D}$ is the network logits before the final classification layer and encodes the uncertainty of one point belonging to each specific semantic class. We define $D = \|\calL\|$ the dimension of the semantic logits and also the number of semantic classes. The semantic logits serve as probabilistic semantic distribution to construct the subsequent Gaussian semantic field. Instances in the upper object layer are further extracted by leveraging clustering algorithms within the instantiable semantic classes, e.g., cars, trunks, and poles. Afterward, the upper semantic object layer of the 3D scene graph is created: $\mathcal{O}=\left\{\mathbf{o}_j,l_j\right\}$, where $\mathbf{o}_j$ is the centroid of the $j$-th instance and $l_j$ is the corresponding semantic label. The instance set is then used for subsequent pose estimation. One may have reservations about selecting moveable objects, e.g., cars, for pose estimation in the global localization scenario. Rather than directly filtering out all such objects, we propose to rerank them by a semantic stability mask as static moveable objects can be good features for data association.

Although the instance layer provides a compact representation for data association and subsequent pose estimation, it suffers from repetitiveness or ambiguity \cite{yin2024outram}. Such a case is illustrated in Fig. \ref{fig: teaser_fig} where a car park lobby is full of repetitive columns. In this semi-symmetric scenario, instance-level semantics can be insufficient to produce reliable data associations for a global optimal. Such scenarios invoke us to use more detailed semantic information for data association. As shown in Fig. \ref{fig: teaser_fig}, the mutual modeling of geometry and background semantics (e.g., vegetation) can be efficient in distinguishing the two repetitive columns apart. As such, we aim to model all semantic classes and their geometric distributions in a unified manner. We introduce an intermediate Gaussian semantic field layer to model this unified geo-semantic distribution of local areas by training a sparse Gaussian process. 

We start from the instance layer by querying the surrounding point cloud around each object centroid $\mathbf{o}_i$ within a fixed length $r$. After the radius search, a group of semantic points is created: $\left\{\mathbf{x}_i, \mathbf{y}_i, l_i \right\}_{i=1}^{M}$ with $\mathbf{x}_i \in \Rbb3$ the 3D location of the point and $\mathbf{y}_i, l_i$ the semantic logits and label respectively. Within the selected point cloud, a multi-layered Gaussian semantic field is constructed to efficiently represent the metric-semantic distribution.

A Gaussian process is a set of random variables such that any finite combination of them is a joint multivariate Gaussian distribution \cite{williams2006gaussian}. We design the Gaussian process to be the mapping between the 3D location in the local coordinate $\mathbf{x}_i$ and the semantic logit $\mathbf{y}_i$ parametrized as $f_{\theta} \colon \mathbb{R}^3 \rightarrow \mathbb{R}^{\|\calL\|}$ where $\theta $ is the hyperparameter set. Assume the latent generative function can be modeled as $\mathbf{y}_i = f_{\theta}(\mathbf{x}_i)+\epsilon_i$, with $\epsilon_i\sim\mathcal{N}(0,\sigma_{y}^2)$ the observation noise.
Given the observation set $\mathcal{D}=\left\{\left(\mathbf{x}_i, \mathbf{y}_i\right)\right\}_{i=1}^{N}$, the GP prior can be written as $f\sim\mathcal{GP}\left(\mu(\mathbf{x}),k(\mathbf{x}, \mathbf{x}^{\prime})\right)$, with $\mu(\mathbf{x})$ the mean function, which is usually set to zero in the absence of prior semantic distribution information, and $k(\mathbf{x}, \mathbf{x}^{\prime})$ the kernel function, which defines the semantic correlation between different positions $\mathbf{x}$ and $\mathbf{x}^{\prime}$. The mean and kernel functions completely describe the Gaussian process \cite{williams2006gaussian}. As is common, a Mat\'ern 3/2 kernel is chosen as the kernel function:
\begin{equation}
\notag
    k(\mathbf x,\mathbf x')
    =\frac{2^{1-\nu}}{\Gamma(\nu)}
    \Bigl(\tfrac{\sqrt{2\nu}\,\lVert \mathbf x-\mathbf x'\rVert}{\kappa}\Bigr)^{\!\nu}
    K_\nu\!\Bigl(\tfrac{\sqrt{2\nu}\,\lVert \mathbf x-\mathbf x'\rVert}{\kappa}\Bigr),
\end{equation}
with $K_{\nu}$ the modified Bessel function, $\nu$ the smoothness, $\kappa$ the length scale of its variation. The predictive mean and covariance of the GP can be derived in closed form by using the Bayes' Rule:
\begin{equation}
\label{eq:gp_mean_variance}
\begin{aligned}
\mu(\mathbf{x}_*) &= k(\mathbf{x}_*, \mathbf{X}) \, \mathbf{K}^{-1} \mathbf{y}, \\
\sigma^2(\mathbf{x}_*) &= k(\mathbf{x}_*, \mathbf{x}_*) 
- k(\mathbf{x}_*, \mathbf{X}) \, \mathbf{K}^{-1} k(\mathbf{X}, \mathbf{x}_*),
\end{aligned}
\end{equation}
where $\mathbf{x}_{*}$ is the 3-D location and $\mathbf{y}$ is the corresponding semantic logits. And $\mathbf{X}$ is the known data point matrix, $\mathbf{K} = k(\mathbf{X}, \mathbf{X}) + \sigma^2 \mathbf{I}$ is the regularized kernel matrix.

\begin{figure}[htb]
\centering
\includegraphics[width=\columnwidth]{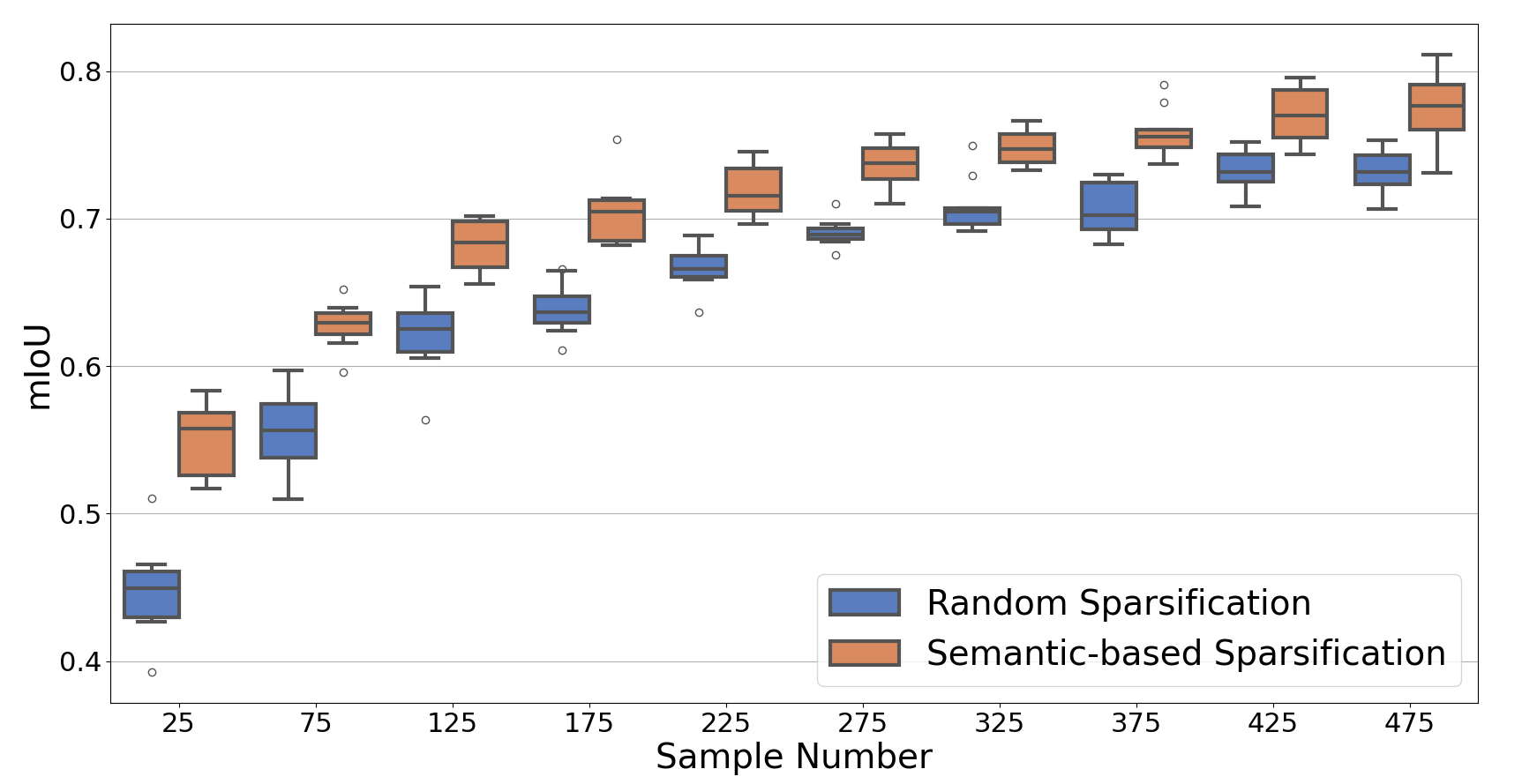}
\caption{Comparison of the semantic reconstruction quality of two downsampling strategies to train the Sparse Gaussian Process in (\ref{eq:gp_mean_variance}). We find that our proposed Semantic-based sparsification consistently outperforms random downsampling in terms of the reconstruction mIoU (mean Intersection over Union) under the same sample number condition.}
\label{fig: abl_sem_spar}
\vspace{-1.5em}
\end{figure}

\myparagraph{Sparse Gaussian Process.} Training a Gaussian process can be computationally intensive when dealing with a large amount of training data with high dimensions. This phenomenon has become increasingly severe considering the density of LiDAR data and the dimension of semantic logits. Works in the literature try to reduce the size of the training set by selecting representative pseudo-supporting points \cite{snelson2005sparse}. In our case, we empirically find the semantic labels to be a natural supporting point selection criterion. More specifically, we construct a semantic class-wise pseudo-training data generation process as illustrated in Algorithm \ref{algo: semantic sparcification}. We find that such a semantic-based sparsification is computationally more efficient and outperforms naive random sparsification. We report the comparison in Fig. \ref{fig: abl_sem_spar}. We find that with the proposed semantic-based sparsification, the trained Gaussian process can better predict the semantic label of the points in the local coverage with higher mIoU.

\begin{algorithm}[h]
\caption{Semantic-based Sparcification}
\label{algo: semantic sparcification}
\KwIn{Point cloud $\mathbf{X} = \{\mathbf{x}_i\}$, Labels $\mathbf{L} = \{\ell_i\}$, Target downsampled point number $N$}
\KwOut{Downsampled point cloud $\mathbf{X'}$, Labels $\mathbf{L'}$}

\BlankLine
\For{each semantic class $c$}{
    $n_c \gets \text{count}(\ell_i = c)$\;
}

\BlankLine
\For{each semantic class $c$}{
    $p_c \gets \frac{n_c}{|\mathbf{X}|}$ \tcp*[l]{Proportion of class $c$}
}

\BlankLine
\For{each semantic class $c$}{
    $N_c \gets \text{round}(p_c \times N)$ \tcp*[l]{Number of points to keep for class $c$}
}

\BlankLine
\For{each semantic class $c$}{
    Randomly select $N_c$ points from $\mathbf{X}$ where $\ell_i = c$\;
}

\BlankLine
$\mathbf{X'}, \mathbf{L'} \gets $ combine all selected points\;

\KwRet $\mathbf{X'}, \mathbf{L'}$\;

\end{algorithm}



\subsection{Triangulated GSF for Substructure Matching}
Given the query scene graph and the reference map, the global localization is achieved in a coarse-to-fine manner. Following \cite{yin2024outram, yuan2023std}, we initiate the localization process by generating sets of triangle descriptors for coarse-level search. We triangulate each scene graph (both the query scan and the map) to form a series of triangles. Given a semantic cluster $\mathcal{A}_i$ in the instance layer of the scene graph, we compute two features from the lower layers: a Gaussian semantic field from the intermediate layer $\mathcal{GSF}(\mathcal{A}_i)$ and a centroid $\mathbf{a}_i$ computed by averaging all instance points in the lower layer. To create the triangle descriptor, each vertex in the object layer $\mathcal{A}_i = \left\{\mathbf{a}_i,\mathcal{GSF}(\mathcal{A}_i)\right\}$ is associated with $K$ nearest clusters $\left\{\mathcal{A}_j\right\}_{j=1}^K$. Afterward, we exhaustively select two of the neighbors, together with the anchor cluster, i.e., $\mathcal{A}_1$, $\mathcal{A}_2$, and $\mathcal{A}_3$, to form one triangle representation of the current scene graph. By an abuse of notation, we denote it as $\Delta\left(\mathcal{A}_{1,2,3}\right)$ which comprises of the following attributes: 
\begin{itemize}
    \item $\mathbf{a}_1$, $\mathbf{a}_2$, $\mathbf{a}_3$: centroids of the semantic clusters;
    \item $\mathcal{GSF}(\mathcal{A}_1), \mathcal{GSF}(\mathcal{A}_2), \mathcal{GSF}(\mathcal{A}_3)$: corresponding Gaussian semantic fields;
    \item $d_{12}$, $d_{23}$, $d_{31}$: three side lengths, $d_{12} \leq d_{23} \leq d_{31}$;
    \item $l_1$, $l_2$, $l_3$: three semantic labels associated with each vertex of the triangle.
\end{itemize}
Similar to STD \cite{yuan2023std} and Outram \cite{yin2024outram}, a hash table is built using the sorted side length $d_{12}$, $d_{23}$, and $d_{31}$ as the key value due to its permutation invariance. Other attributes are left for verification purposes. In the searching process, we have the triangulated scene graph in the query scan and reference map:
\begin{equation}
\begin{aligned}
    \Delta{\textit{Query}} =\left\{\Delta\left(\mathcal{A}_{1,2,3}^n\right)\right\}_{n = 1}^{N}, \\
    \Delta{\textit{Map}} =\left\{\Delta\left(\mathcal{B}_{1,2,3}^m\right)\right\}_{m=1}^{M},
\end{aligned} 
\end{equation}
where $n$ and $m$ are indices for triangle descriptors in the query and map scene graph, respectively. We drop the subscript and denote $\Delta\left(\mathcal{A}_{1,2,3}^n\right)$ as $\Delta\mathcal{A}^n$ for clarity. Querying each of the triangles (e.g., $\Delta\mathcal{A}^1$) against the hash table constructed by the reference semantic scene graph will produce multiple responses $\left\{\Delta\mathcal{B}^q\right\}_{q = 1}^{Q}$ as similar substructures could exist throughout the whole mapping region. 

After the first round of coarse matching, we leverage the GSF for fine-level filtering. While the Gaussian process itself perfectly models the joint metric semantic distribution of areas of interest, it is typically hard to use it directly as a metric-level similarity measurement \cite{mallasto2017learning}. We lend the trained Gaussian semantic field the ability for similarity comparison by a sampling process called grid probing. Given a trained Gaussian process $\mathcal{GP}$, a set of responses can be produced by querying the process with inputs, where in our case is a set of 3D locations.

\myparagraph{Similarity Measurement for Gaussian processes.}
While Gaussian processes are non-parametric representations, there exists a 2-Wasserstein metric for them \cite{mallasto2017learning} achieved by sampling the Gaussian measure and computing the 2-Wasserstein between the multivariate Gaussian distributions accordingly. Define the sampling supports $\mathcal{S}$ as a uniform 2D grid:
\begin{align*}
\mathcal{S} = \big\{ (x_i, y_j) \;\big|\; 
& x_i = x_0 + i \cdot \Delta x,\; 0 \leq i < N_x, \\
& y_j = y_0 + j \cdot \Delta y,\; 0 \leq j < N_y 
\big\},
\end{align*}
where $(x_0, y_0)$ denotes the centroid of one semantic cluster, $\Delta x$ and $\Delta y$ are the sampling intervals, and $N_x$, $N_y$ are the number of grid points along each axis. The sampling result is a group of multivariate Gaussians defined as the populations of the GP: $ \mathcal{N}\left(\boldsymbol{\mu},\, \mathbf{K} + \sigma^2 \mathbf{I} \right)$. The distance between two Semantic distance fields can be defined using the 2-Wasserstein distance between the populations of GPs:
\begin{equation}
\begin{aligned}
W_2^2\left( \mathcal{GSF}(\mathcal{A}_1),\, \mathcal{GSF}(\mathcal{B}_1) \right) 
& := \;  \| \boldsymbol{\mu}_1 - \boldsymbol{\mu}_2 \|_2^2 \\
& 
+ \operatorname{Tr}(\boldsymbol{\Sigma}_1 + \boldsymbol{\Sigma}_2) \\
& 
- 2\, \operatorname{Tr}\left( 
\left( \boldsymbol{\Sigma}_1^{1/2} \boldsymbol{\Sigma}_2 \boldsymbol{\Sigma}_1^{1/2} \right)^{1/2}
\right),
\end{aligned}
\label{eq:wasserstein}
\end{equation}
with $\mathbf{\Sigma}_1 = \mathbf{K}_1 + \sigma_{1}^2 \mathbf{I}$ and $\mathbf{\Sigma}_2 = \mathbf{K}_2 + \sigma_{2}^2 \mathbf{I}$. 

\myparagraph{Semantic Stability Mask.} Inspired by \cite{chen2019suma++, xue2023sfd2}, a stability mask is proposed to tackle the possible scene changes. Explicitly, instead of directly filtering out movable semantic classes from segmentation results, we empirically assign each semantic class a stability value $w_i$ according to its possibility to move and divide the semantic classes into types of volatile, short-term, and long-term. Computationally, this is achieved by alternating the covariance matrix $\Sigma$ with a semantic-stability-weighted version $\mathbf{W}^{1/2} \boldsymbol{\Sigma} \mathbf{W}^{1/2}$, with $\mathbf{W}=diag(w_1,w_2,...,w_l)$. The weighted version can be viewed as the semantic-based Mahalanobis distance.

\begin{table}[t]
\caption{Semantic Stability Table}
    \begin{center}
    \begin{tabular}{ccccc}
    \toprule
    Sem. Class & Volatile & S-Term & L-Term & Sta. Value $w_i$ \\
    \midrule
    Truck, Bike, Person & \checkmark & & & 0.1\\
    Car, Nature & & \checkmark & & 0.5\\
    Infra. & & & \checkmark & 1.0\\
    \bottomrule
    \end{tabular}
    \end{center}
    \label{tab: dataset stats}
    \vspace{-2.5em}
\end{table}

With the help of the proposed Gaussian semantic field, a more detailed semantic distribution is considered in a continuous manner.
Following the exhaustive enumeration of triangular structures within the query scene graph, we construct an initial collection of instance-level correspondences $\mathcal{I}_\text{raw}$ by exploiting the inherent ordering of edge lengths, which establishes a natural association mechanism between semantic cluster pairs.

\subsection{Global Localization}
With the raw GSF-wise correspondence $\mathcal{I}_\text{raw}$ built in the above section, we employ the maximum clique process to find out the inlier correspondences. Intuitively, we seek to find an area that maximizes the number of mutually consistent correspondences as well as maintains the consistency between these local structures:
\begin{align}
\begin{split}
    &\mathop{\max}_{\mathcal{I}\subset\mathcal{I}_\text{raw}} \quad \left|\mathcal{I}\right| \\ 
    &\text{s.t.}\ \mathcal{D}\left(\mathcal{I}_i, \mathcal{I}_j\right) \leq\epsilon,\ \forall \mathcal{I}_i, \mathcal{I}_j\in \mathcal{I},
    \label{equ: max inlier ch5}
\end{split}
\end{align}
with $\mathcal{D}$ the metric consistency check which indicates whether two correspondences are mutually consistent with each other and $\epsilon$ the threshold. Namely, for two correspondences $\mathcal{I}_i$ and $\mathcal{I}_j$, with their corresponding semantic clusters $\mathcal{A}_i, \mathcal{B}_i$ and $\mathcal{A}_j, \mathcal{B}_j$, a consistency check is defined as
\begin{equation}
    \mathcal{D}\left(\mathcal{I}_i, \mathcal{I}_j\right) \triangleq \text{dist}\left(\mathcal{A}_{ij}, \mathcal{B}_{ij}\right).
\end{equation}
We define $\mathcal{A}_{ij}:=\mathcal{A}_{i} - \mathcal{A}_{j}$ and $\mathcal{B}_{ij}:=\mathcal{B}_{i} - \mathcal{B}_{j}$ as the Euclidean distance between the corresponding two vertices. The consistency check defines how two correspondences $\mathcal{D}\left(\mathcal{I}_i, \mathcal{I}_j\right)$ agree with each other. We solve the objective function \ref{equ: max inlier ch5} by parallel maximum clique \cite{rossi2015parallel}.


Having established continuous semantic field correspondences through our GSF framework, we proceed to estimate the global pose by leveraging the weighted semantic field similarities. Unlike traditional discrete matching approaches, our continuous field representation provides probabilistic correspondence confidences that naturally integrate into the optimization process.

Given the set of semantic field correspondences with their associated confidence scores from the GSF similarity computation, we formulate the global localization as a weighted least squares optimization problem:
\begin{equation}
    \hat{\mathbf{R}},\hat{\mathbf{t}}=\mathop{\arg\min}_{{\bf R}\in SO(3), {\bf t}\in \mathbb{R}^3}\sum\limits_{ij\in\mathcal{C}}\omega_{ij}\left\|\mathbf{p}_i-\mathbf{R}\mathbf{q}_j-\mathbf{t}\right\|_2^2,
    \label{eq: gsf_localization}
\end{equation}
where $\mathcal{C}$ represents the set o
f continuous semantic field correspondences, and $\omega_{ij}$ denotes the confidence weight derived from the GSF similarity measurement between semantic field patches $i$ and $j$ from (\ref{eq:wasserstein}). The confidence weights $\omega_{ij}$ are computed from the continuous semantic field similarities, automatically emphasizing correspondences with high semantic consistency while down-weighting ambiguous matches.

To handle potential outliers that may arise from semantic ambiguities in highly repetitive scenes, we incorporate a robust truncated formulation:
\begin{equation}
    \hat{\mathbf{R}},\hat{\mathbf{t}}=\mathop{\arg\min}_{{\bf R}\in SO(3), {\bf t}\in \mathbb{R}^3}\sum\limits_{ij\in\mathcal{C}}\omega_{ij}\min\left(\left\|\mathbf{p}_i-\mathbf{R}\mathbf{q}_j-\mathbf{t}\right\|_2^2,\tau_{ij}\right),
    \label{eq: robust_gsf_localization}
\end{equation}
where $\tau_{ij}$ represents an adaptive truncation threshold based on the semantic field confidence. This formulation naturally combines the benefits of continuous semantic field disambiguation with robust geometric optimization, enabling accurate localization even in challenging repetitive environments.

\begin{table*}[t]

\caption{Global Localization Performance Comparison}
    \vspace{-1em}
    \begin{center}
        
    \begin{tabular}{clcccccccccc}
    \toprule
    & & \multicolumn{9}{c}{Successful Global Localization Rate [\%] $\uparrow$} \\
    \cmidrule(lr){3-12}
    & Dataset & \multicolumn{2}{c}{MulRan DCC} & \multicolumn{2}{c}{MulRan KAIST} & \multicolumn{6}{c}{MCD NTU} \\
    \cmidrule(lr){3-4}
    \cmidrule(lr){5-6}
    \cmidrule(lr){7-12}
    & Localization Seq. & 01 & 02 & 01 & 03 & 01 & 02 & 04 & 10 & 13 & repet.\\
    \midrule
    
    \multirow{5}{*}{\rotatebox{90}{LCD}} & GOSMatch \cite{zhu2020gosmatch} & 48.61 & 50.17 & 35.93 & 51.98 & 50.01 & 52.72 & 46.83 & 53.31 & 51.32 & 30.06 \\
    
    & STD \cite{yuan2023std} & 17.57 & 18.06 & 49.61 & 38.96 & 42.54 & 53.78 & 67.52 & 52.65 & 61.89 & 32.16\\

    & BTC \cite{yuan2024btc} & 20.08 & 32.25 & 52.97 & 65.17 & 46.72 & 55.85 & 67.23 & 52.47 & 68.12 & 34.08 \\

    & BEVplaces++ \cite{luo2024bevplace++} & 46.67 & 51.47 & 86.85 & \underline{90.15} & 71.12 & 80.31 & 69.48 & 73.53 & 82.72 & 52.15\\

    & Ring++ \cite{xu2023ring++} & \underline{89.54} & \underline{91.67} & 85.42 & 88.97 & \underline{93.78} & \underline{96.45} & 95.08 & \underline{97.12} & 96.36 & \underline{60.05}\\
    
    \midrule
    
    \multirow{5}{*}{\rotatebox{90}{Regis.}} & Ankenbauer et al. (Original) \cite{ankenbauer2023global} & - & - & - & - & 75.32 & 78.01 & 80.42 & 72.84 & 83.35 & 45.89\\
    
    & Ankenbauer et al. (Cons.) \cite{ankenbauer2023global} & 0.072 & 0.032 & 0.025 & 0.012 & - & - & - & - & - & - \\
    

    & SGTD \cite{huang2025sgtd} & 81.15 & 92.37 & \underline{86.87} & 92.06 & 93.65 & 92.31 & 95.26 & 96.51 & 96.27 & 55.26 \\
    
    & Outram \cite{yin2024outram} & 82.53 & 90.48 & 84.41 & 85.64 & 92.11 & 93.79 & \underline{95.51} & 96.01 & \underline{96.73} & 54.72 \\

    & \textbf{Outram-GSF (Ours)} & \textbf{90.83} & \textbf{94.27} & \textbf{91.09} & \textbf{93.26} & \textbf{98.51} & \textbf{98.43} & \textbf{96.76} & \textbf{99.32} & \textbf{98.53} & \textbf{87.36}\\
    
    \bottomrule
    \end{tabular}
    \end{center}
    \label{tab: success rate}
\end{table*}

We solve Eq.~(\ref{eq: robust_gsf_localization}) using an iterative reweighted least squares approach that alternates between updating the pose estimate and refining the semantic field correspondences based on geometric consistency.

\section{Experimental Results}

\label{sec: results}
In this section, we compare our proposed method with state-of-the-art LiDAR-based one-shot global localization methods. For LCD-based methods, we involve handcrafted sota descriptors BTC \cite{yuan2024btc} and learning-based sota BEVplaces++ \cite{luo2024bevplace++} and Ring++ \cite{xu2023ring++}. For registration-based baselines, apart from previous baselines in Outram \cite{ankenbauer2023global, yin2024outram}, a newly proposed SOTA SGDT \cite{wang2025sgt} is included. All mentioned algorithms are tested on a PC with Intel i9-13900 and 32Gb RAM with Nvidia RTX4080.

\myparagraph{Experiment Setup.} We evaluate our proposed method, Outram-GSF, against several state-of-the-art on two publicly available datasets: MulRan \cite{kim2020mulran} and MCD \cite{nguyen2024mcd}. To further investigate the performance of different algorithms in highly repetitive environments, we derived a sequence out of MCD NTU seq01 featuring highly semantic symmetry. Furthermore, to mimic a real global localization or relocalization scenario, different from a loop closure detection setting, we intentionally involve temporal diversity between the mapping or descriptor generation session and the localization session from days to months. For each mapping sequence, we concatenate semantically annotated scans \cite{cao2023multi} by the ground truth pose to generate the semantic segmented reference map for registration-based methods. All semantic classes with instance definitions are incorporated for localization. For LCD-based global localization methods, frames in the mapping sequences are encoded into a database for retrieval using scans in the localization sequence. 
Statistics of the benchmark datasets are presented in Table \ref{tab: dataset stats}. 
The criteria for choosing the mapping sequence are the sequence that has the most coverage of the target area.

\subsection{Localization Success Rate}

\begin{table}[t]
\caption{Details of Evaluation Datasets}
    \begin{center}
    \begin{tabular}{c|cc|c}
    \toprule
    Mapping/Loc. Sequence & Length & Scan Number & Time Diff. \\
    \midrule
    \textit{Mapping:} \\
    MulRan DCC 03   & 5.7 km & 7479 & - \\
    MulRan KAIST 02 & 6.3 km & 8941 & - \\
    MCD NTU 08      & 3.8 km & 6023 & - \\
    \midrule
    \textit{Localization:}\\
    MulRan DCC 01   & 4.9 km   & 5542 & 20 days\\
    MulRan DCC 02   & 5.2 km   & 7561 & 1 month\\
    MulRan KAIST 01 & 6.3 km   & 8226 & 2 months\\
    MulRan KAIST 03 & 6.4 km   & 8629 & 10 days\\
    MCD NTU 01      & 3.8 km & 6023 & 2 hours \\
    MCD NTU 02      & 0.64 km  & 2288 & 2 hours\\
    MCD NTU 04      & 0.64 km  & 2288 & 2 hours\\
    MCD NTU 10      & 0.64 km  & 2288 & 2 hours\\
    MCD NTU 13      & 1.23 km & 2337  & 2 days\\
    MCD NTU 01 \textit{repet.}  & 0.26 km & 572  & 2 hours\\
    \bottomrule
    \end{tabular}
    \end{center}
    \label{tab: dataset stats}
\end{table}

We adopt the same evaluation metrics as in prior works \cite{yin2024outram,huang2025sgtd}. The principal evaluation metric in global localization is the success rate of localization, where every frame in the localization sequence will be queried against a map built by a mapping sequence. As demonstrated in Table~\ref{tab: success rate}, our proposed method, Outram-GSF, establishes consistent performance advantages over all other counterparts.
On MulRan DCC, our method achieves 90.83\% and 94.27\% on sequences 01 and 02, respectively, outperforming previous best methods including SGTD \cite{huang2025sgtd} and Outram \cite{yin2024outram}. On MulRan KAIST, Outram-GSF again leads with up to 93.26\% localization success, demonstrating strong generalization across different urban environments. Notably, on MCD NTU, our method reaches 99.32\% on sequence 10 and maintains robust performance across all sequences, with an overall repetitive-scene success rate of 87.36\%, surpassing SGTD (55.26\%) and Outram (54.72\%) by a significant margin.

Compared to LCD-based approaches, which tend to struggle under appearance changes and repetitive structures, Outram-GSF shows substantial improvements. While Ring++ \cite{xu2023ring++} performs best among LCD methods, its repetitive-scene performance (60.05\%) still lags behind our method by over 27\%. Other LCD baselines such as GOSMatch \cite{zhu2020gosmatch}, STD \cite{yuan2023std}, and BTC \cite{yuan2024btc} perform noticeably worse, particularly in challenging scenarios.

These results highlight the superiority of Outram-GSF in achieving robust and accurate global localization across diverse and challenging environments. The gains over prior registration-based methods confirm the effectiveness of our framework in resolving long-range data associations and handling semantic aliasing, making it well-suited for real-world deployments. More importantly, our method performs noticeably better on the sequence featuring repetitive scenes, compared to previous semantic registration-based sota methods \cite{yin2024outram, huang2025sgtd}. This observation validates the effectiveness of the Gaussian semantic field in modeling fine-grained geo-semantic information for semantic disambiguation. It is also worth noting that the proposed Outram-GSF works without any post-geometric verification, whereas it outperforms verification-based counterparts \cite{huang2025sgtd, yuan2024btc}.

\begin{table}[t]
\vspace{5pt}
\caption{Average Translation/Rotation Error and Runtime}
    \begin{center}
    \begin{tabular}{ccccc}
    \toprule
    & \multicolumn{3}{c}{ATE/ARE [\text{meter}/\degree]} & \multirow{2}{*}{t [ms]}\\
    \cmidrule(lr){2-4}
    & MulRan DCC & MulRan Kaist & MCD NTU \\
    \midrule
    STD & 2.09/1.52 & 0.68/0.55 & 0.56/1.67 & \textbf{9.34}\\
    BTC & \underline{1.41}/1.56 & \textbf{0.53}/0.76 & 0.62/1.42 & \underline{13.2}\\
    BEVplaces++ & 2.76/1.95 & 1.04/1.68 & 1.27/2.05 & 227.6\\
    Ring++ & 1.62/0.75 & 0.63/\underline{0.50} & 0.48/\textbf{1.15} & 382.6\\
    SGDT & 1.51/\underline{0.59} & \underline{0.54}/0.52 & \textbf{0.36}/1.28 & 123.3 \\
    Outram & 1.55/1.82 & 0.92/0.83 & 0.50/1.86 & 305.7 \\
    Outram-GSF & 1.54/1.88 & 0.87/0.72 & 0.60/2.01 &  532.7\\
    Ours w/ rf & \textbf{1.38}/\textbf{0.52} & 0.55/\textbf{0.48} & \underline{0.41}/\underline{1.28} & 563.6\\
    
    \bottomrule
    \end{tabular}
    \end{center}

\end{table}

\subsection{Pose Estimation Accuracy and Runtime Results}

We evaluate the localization accuracy and runtime efficiency of our proposed method on all sequences of the three datasets: MulRan DCC, MulRan KAIST, and MCD NTU. A variant of our proposed method with subsequent GICP \cite{koide2024small_gicp} refinement (Ours w/ rf) is also presented. As reported in Table IV, our method with refinement (Ours w/ rf) achieves the lowest average translation and rotation errors across all benchmarks, demonstrating superior pose estimation accuracy under diverse urban conditions.

Specifically, our method achieves an average translation error (ATE) of 1.38m, 0.55m, and 0.41m, and a rotation error (ARE) of 0.52\degree, 0.48\degree, and 1.28\degree on MulRan DCC, KAIST, and MCD NTU, respectively. These results represent state-of-the-art performance, significantly outperforming both retrieval-based methods (e.g., STD, BEVplaces++) and other registration-based baselines (e.g., SGTD, Outram).

While our refined pipeline incurs higher computational cost (563.6 ms/frame), the substantial gain in accuracy justifies the trade-off for applications that prioritize robust and precise global localization. Compared to fast but less accurate methods such as STD \cite{yuan2023std} (9.34 ms/frame) and BTC \cite{yuan2024btc}, our approach is better suited for long-range, large-scale deployment where localization reliability is essential.

These results confirm the effectiveness of our global registration framework employing the Gaussian semantic field in delivering high-fidelity 6-DoF pose estimates across complex real-world environments.

\begin{figure}[htb]
\centering
\includegraphics[width=\columnwidth]{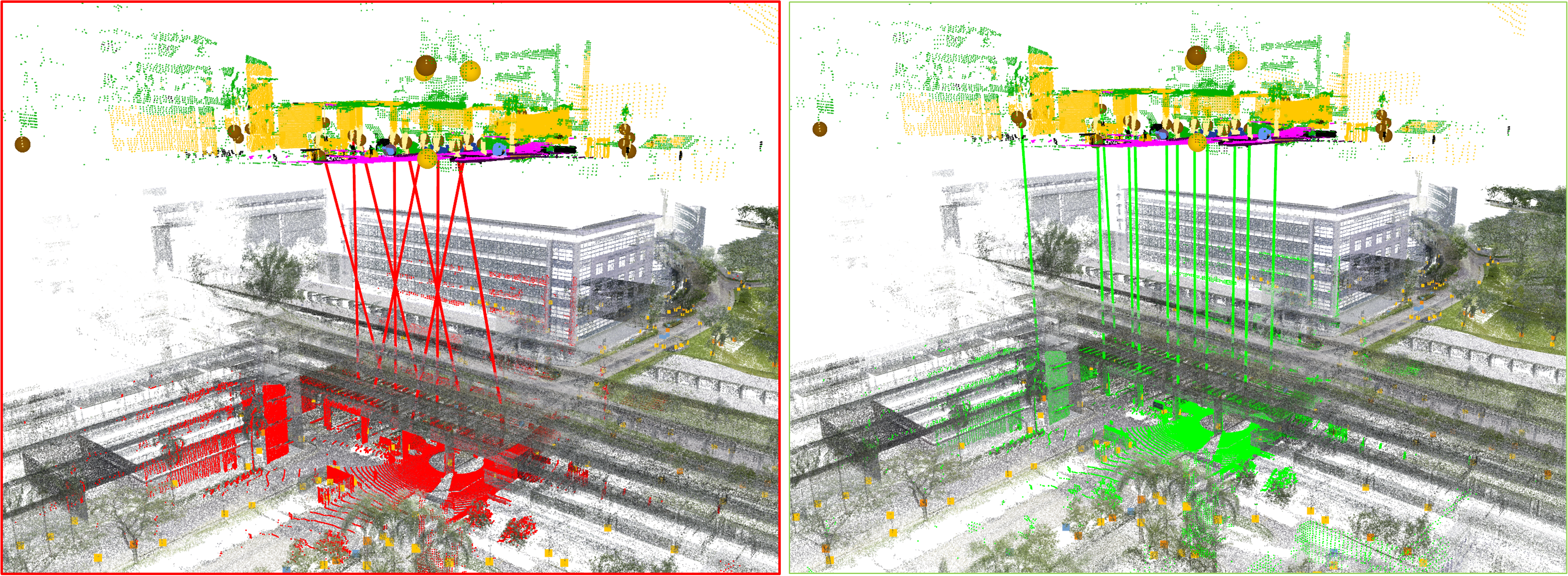}
\caption{A visual comparison of the correspondence establishment and global localization performance in a symmetric environment of the proposed method with (right) and without (left) the Gaussian semantic field. Pure instance-level methods result in twisted correspondences (see the reversal correspondences in red
).}
\label{fig: viz_cmp}
\end{figure}

\section{Limitations}
Although our proposed Outram-GSF demonstrates robust performance across multiple datasets encompassing diverse environmental conditions, it may encounter limitations in certain scenarios, particularly in purely symmetric or highly repetitive environments. This observation motivates the development of symmetry-aware algorithms that can assess and provide localizability certificates for given environments, thereby predicting the feasibility of reliable localization prior to deployment.

\section{Conclusion} 
\label{sec:conclusion}
This work introduces Outram-GSF, a one-shot LiDAR-based global localization framework. Unlike previous approaches in the field, our method employs a Gaussian semantic field representation to capture continuous semantic distributions, moving beyond the limitations of conventional cluster centroid-based techniques. The resulting localization system demonstrates exceptional robustness and achieves superior performance compared to state-of-the-art methods across diverse benchmark datasets, particularly excelling in challenging scenarios characterized by geometric repetition and structural symmetry.

\section*{Acknowledgments}



{
\bibliographystyle{IEEEtran}
\bibliography{references}
}

\end{document}